\documentclass[11pt]{article}
\usepackage[margin=1in]{geometry}
\usepackage[T1]{fontenc}
\usepackage{lmodern}
\usepackage{amsmath,amssymb}
\usepackage{booktabs}
\usepackage{array}
\usepackage{graphicx}
\usepackage{xcolor}
\usepackage{tikz}
\usetikzlibrary{decorations.pathreplacing}
\usepackage[numbers,sort&compress]{natbib}
\usepackage[hidelinks]{hyperref}
\usepackage{microtype}
\newcommand{\R}{\mathcal{R}}
\newcommand{\U}{\mathcal{U}}
\newcommand{\rope}{\operatorname{RoPE}}
\newcommand{\rms}{\operatorname{RMSNorm}}

\title{Storing the Gemma~4 K=V Cache Once:\\Exact Global-KV Memory Savings in a Paged Serving Engine\\with Unchanged Attention Kernels}
\author{Kushagra Bainsla\\San Jos\'e State University\\\texttt{kushagra.bainsla@sjsu.edu}}
\date{November 2026}

\begin{document}
\maketitle

\begin{abstract}
In the global attention layers of Gemma~4 31B, the value is the normalized key projection, and the key is the value scaled by a fixed weight, with rotary position encoding on one quarter of its dimensions. The Gemma~4 report states that this design reduces the global KV cache by up to 37.5\%, but the open serving engines that we checked still store a full key and a full value: 1{,}024 values per KV head and token. We store one 640-value row instead, exact in real arithmetic, and an unchanged paged attention kernel reads it through two overlapping strided views as its key cache and its value cache. In a vLLM prototype with the 16-bit 31B model, the outputs are bit-identical to an emulation of the same arithmetic on the stock cache, 2 of 432 final answers differ from the stock engine (within its run-to-run noise), and the steady decode KV memory per request falls by a factor of 1.34 to 1.40 at 32K to 64K tokens. On a prefill/decode pair of two A100 GPUs with NIXL transfer and CPU offload, a decode engine with a full 15~GiB pool ran 1.38 times the stock decode tokens per step for 16K-token prompts (repeated to 0.06\%) and 1.24 times for 8K-token prompts; a model of the vLLM block pool predicted both ratios to within 1\% (the engine was 0.7\% to 0.9\% higher) and matched the change between them. The eager decode throughput rose by 1.30 and 1.11 times, and on this path the compact engine gave the same output tokens as the emulation for 8 prompts at batch 1. The run-time prototype makes an eager step 4\% to 9\% slower at equal batch, and the method does not yet stack with an 8-bit KV cache at the default scale.
\end{abstract}

\section{Introduction}

The KV cache sets the number of long requests that one GPU can serve. Paged serving engines such as vLLM~\cite{vllm} keep the cache in fixed-size blocks and admit a request only when its blocks fit in the pool. A smaller cache per request therefore allows a larger batch and more decode tokens per second, until the GPU becomes compute bound.

Gemma~4~\cite{gemma4} uses two layer types. In the 31B model, 50 sliding-window layers attend to the last 1{,}024 tokens with 16 KV heads of 256 dimensions, and 10 global layers attend to the whole context with 4 KV heads of 512 dimensions. At long context the global layers hold most of the cache of a request. They also have a special structure, which the model configuration calls \texttt{attention\_k\_eq\_v}: they have no value projection. The value is the RMS-normalized key projection, and the key is that value multiplied by the key-norm weight, with rotary position encoding (RoPE)~\cite{rope} on 128 of the 512 dimensions. The Gemma~4 report states that this design reduces the global KV cache by up to 37.5\%~\cite{gemma4}. The 26B-A4B and 12B models use the same design; the E2B and E4B models do not.

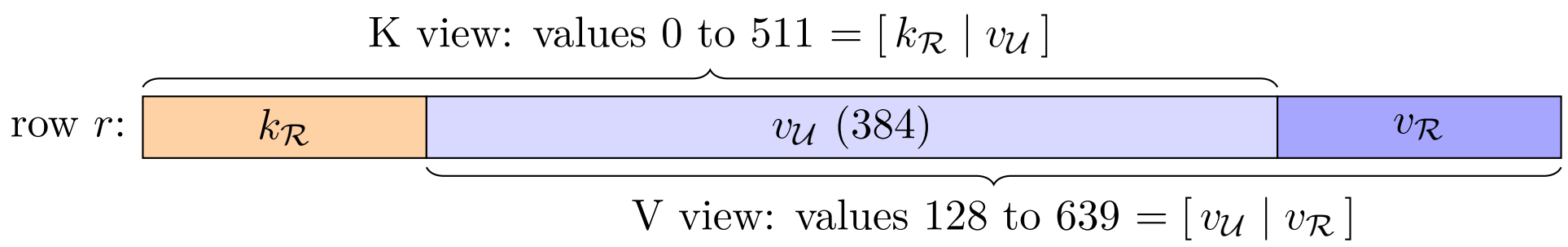
\begin{figure}[t]
\centering
\begin{tikzpicture}[x=0.018cm,y=0.5cm,font=\small]
\draw[fill=orange!35] (0,0) rectangle (128,1) node[midway] {$k_\R$};
\draw[fill=blue!15] (128,0) rectangle (512,1) node[midway] {$v_\U$ (384)};
\draw[fill=blue!35] (512,0) rectangle (640,1) node[midway] {$v_\R$};
\node[left] at (0,0.5) {row $r$:};
\draw[decorate,decoration={brace,amplitude=4pt}] (0,1.15) -- (512,1.15) node[midway,above=3pt] {K view: values 0 to 511 $=[\,k_\R \mid v_\U\,]$};
\draw[decorate,decoration={brace,amplitude=4pt,mirror}] (128,-0.15) -- (640,-0.15) node[midway,below=3pt] {V view: values 128 to 639 $=[\,v_\U \mid v_\R\,]$};
\end{tikzpicture}
\caption{The compact row of one global KV head and token (640 values instead of 1{,}024). A stock attention kernel reads the K view and the V view as its two caches. The query carries $w_\U$ on its unrotated part (Eq.~\eqref{eq:fold}), and the output is permuted back to the model order.}
\label{fig:row}
\end{figure}

The open serving engines that we checked do not realize this saving: they store 1{,}024 values per KV head and token in these layers (Section~\ref{sec:related}). This paper shows how to store 640 values that an unchanged paged attention kernel can read, and it measures what this gives in vLLM. The kernel reads the 640-value row through two views: a view is a window on the same memory with its own shape and strides, so no data is copied (Figure~\ref{fig:row}).

We do not claim the saving, the shape of the stored row or the exact cache reduction as new. The Gemma~4 report states the saving. Multi-head latent attention (MLA)~\cite{deepseekv2} caches a shared part plus a separate RoPE part, and its decode kernels read the key and the value from one cached row~\cite{flashmla}. Slim attention~\cite{slimattention} caches only the keys of multi-head attention layers and computes the values from them with a matrix product. Our two-view read is the MLA decode layout specialized to equal key and value widths, so that kernels that require equal key and value head sizes read the shared buffer unchanged; our searches targeted Gemma~4 and its engines, not this layout in general. Our contributions are:
\begin{enumerate}
\item \textbf{A compact row and a two-view read.} One row of 640 values per KV head and token holds the rotated part of the key and the whole value. A stock paged attention kernel reads it through two overlapping strided views; the key-norm weight folds into the query, and a fixed permutation restores the output order. The method is exact in real arithmetic (Section~\ref{sec:method}).
\item \textbf{A checked vLLM prototype.} Its outputs are bit-identical to an emulation of the same arithmetic on the stock cache, and its final answers stay within noise of the stock engine (Sections~\ref{sec:impl} and~\ref{sec:accuracy}).
\item \textbf{Measurements and a speed bound.} We measure the memory per request, the batch that a fixed pool holds and the step cost of the prototype. A byte count shows why the smaller cache gives little speed per step at these batch sizes: the weights are about 85\% of the bytes that a decode step reads (Sections~\ref{sec:capacity} to~\ref{sec:bytes}).
\item \textbf{A test in a serving decode node.} The compact cache works with the vLLM CPU offload connector and, with a small patch for the pinned build, with NIXL transfer from a prefill engine. On a two-GPU prefill/decode pair whose decode pool stays full, the compact decode engine ran 1.38 and 1.24 times the stock decode tokens per step at two operating points. A model of the block pool, frozen before each run, predicted both ratios to within 1\% (the engine was 0.7\% to 0.9\% higher) and matched the change between them. The eager decode throughput rose by 1.30 and 1.11 times, and the compact engine gave the same output tokens as the emulation for 8 prompts at batch 1 (Section~\ref{sec:node}).
\end{enumerate}

An 8-bit (FP8) KV cache halves all KV bytes, but it is lossy. We tested the compact row with FP8 only in attention tests (Appendix~\ref{app:kernels}), so we do not claim that the two savings stack.

\section{Background}
\label{sec:background}

\paragraph{Gemma~4 global attention.} Let $x$ be the input of a global layer and $y = W_k x$ the key projection of one KV head ($y \in \mathbb{R}^{512}$). The layer has no $W_v$. It computes
\begin{equation}
v = \rms(y), \qquad k = \rope_p\!\left(w \odot v\right),
\label{eq:kv}
\end{equation}
where $\rms$ has no learned scale for $v$. Here $w \in \mathbb{R}^{512}$ is the weight of the key norm as the model applies it (one vector per layer, shared by all KV heads), and $\rope_p$ rotates a fraction $p = 0.25$ of the dimensions (p-RoPE~\cite{prope}). RoPE pairs dimension $i$ with dimension $i + 256$, and only 64 of the 256 pairs have a nonzero frequency. So the rotated set is $\R = \{0, \dots, 63\} \cup \{256, \dots, 319\}$ ($|\R| = 128$), and the unrotated set $\U$ holds the other 384 dimensions. In words: one quarter of the dimensions carry the token position, and three quarters do not. On $\U$, RoPE is the identity, so
\begin{equation}
k_\U = w_\U \odot v_\U, \qquad k_\R = \rope\!\left(w_\R \odot v_\R\right).
\label{eq:split}
\end{equation}
The 31B model has 32 query heads, so 8 query heads share each global KV head. The query is $q = \rope_p(\mathrm{qnorm}(W_q x))$, where qnorm is the RMS norm of the query. The global layers use an attention scale of 1.0 and no soft-cap on the attention logits.

\paragraph{Paged KV caches.} vLLM~\cite{vllm} keeps the KV cache in a fixed pool of GPU memory, which it splits into blocks of $B$ tokens; a block holds one page in each layer of a group. For a hybrid model, the vLLM version that we used~\cite{vllmcode} splits the layers into groups of equal size; for Gemma~4 31B these are 5 sliding groups and 1 global group, with 10 layers each. All groups must use one page size in bytes. The code that enforces this, which we call the page unifier, grows the block size of a group whose page divides the largest page, and otherwise pads the smaller page. The scheduler admits a waiting request, that is, starts its prefill, only if its blocks fit: at this commit the option \texttt{scheduler\_reserve\_full\_isl} (ISL: input sequence length) is on by default, so the full prefill reservation must fit in the free blocks. If a running request needs a block and none is free, the engine preempts a request: it stops that request and computes it again later. The Triton~\cite{triton} attention backend of vLLM takes the key cache and the value cache as two tensors with explicit strides, and it writes new keys and values with a separate cache-write kernel.

\paragraph{Eager mode and CUDA graphs.} In eager mode, the engine launches each GPU kernel of a decode step from Python, so many small kernels cost CPU time. With CUDA graphs, it records the kernels of a decode step once and replays them with one launch.

\section{Method}
\label{sec:method}

\paragraph{What to store.} By Eq.~\eqref{eq:split}, the pair $(k, v)$ of a head and token is a function of $(v, k_\R)$: $k_\U$ is $w_\U \odot v_\U$, and $w$ is a fixed model weight. So 512 + 128 = 640 values hold what the stock cache stores in 1{,}024 values, which is 37.5\% fewer values.

\paragraph{Why not 512 values.} Since $k_\R = \rope(w_\R \odot v_\R)$, the value and the token position determine the whole key, so 512 values would be enough (50\% instead of 37.5\%). But the rotation angle differs for each cached token, so it cannot be applied once to the query; a 512-value row needs a new kernel that rotates $w_\R \odot v_\R$ for each token while it reads the cache. We keep the 128 values of $k_\R$ so that an unchanged kernel can read them.

\paragraph{Scores without $k_\U$.} For a query $q$ of a head that reads this KV head,
\begin{equation}
q \cdot k = q_\R \cdot k_\R + q_\U \cdot (w_\U \odot v_\U) = q_\R \cdot k_\R + (q_\U \odot w_\U) \cdot v_\U .
\label{eq:fold}
\end{equation}
So with the folded query $q' = [\,q_\R \mid q_\U \odot w_\U\,]$, the key can be read as $[\,k_\R \mid v_\U\,]$. The fold is exact in real arithmetic for any $w$, because RoPE does not touch $\U$.

\paragraph{The row and its two views.} We store, per KV head and token, the row
\begin{equation}
r = [\,k_\R\ (128) \mid v_\U\ (384) \mid v_\R\ (128)\,] \in \mathbb{R}^{640}.
\end{equation}
Its first 512 values, $[\,k_\R \mid v_\U\,]$, are the key that Eq.~\eqref{eq:fold} needs (the \emph{K view}). Its last 512 values, $[\,v_\U \mid v_\R\,]$, are the value in a permuted dimension order (the \emph{V view}). Figure~\ref{fig:row} shows the layout. Both views are strided tensors over the same buffer, with the same block, token and head strides and start offsets of 0 and 128 values, so a kernel that takes separate key and value caches with strides reads them unchanged. The attention output comes in the V-view order $[\,\U \mid \R\,]$, and a fixed permutation restores the model order before the output projection.

\paragraph{Writing.} The stock cache-write operation stores a key and a value per token. One call fills the row: we give it $k' = [\,k_\R \mid v_\U\,]$ as the key, for the K view, and $v' = [\,v_\U \mid v_\R\,]$ as the value, for the V view. The key write and the value write put the same values into the shared region $v_\U$, so their order does not matter. With an FP8 cache, the two writes must use one scale for the shared region, so the key scale and the value scale must be tied, per layer or per row (Appendix~\ref{app:kernels}).

\paragraph{Exactness.} In real arithmetic the scores and the outputs equal those of the stock cache. In 16-bit arithmetic the method is not bit-exact. Let $n$ be the normalized key projection in 32-bit arithmetic. The stock cache stores $k_\U$ as one rounding of $w_\U \odot n_\U$ and $v_\U$ as one rounding of $n_\U$; the compact score uses the rounded $v_\U$ with a rounded $q_\U \odot w_\U$ instead, so the roundings differ. The permuted order of the dimensions also changes the order of the floating-point sums. Neither layout is the real-arithmetic reference. Section~\ref{sec:accuracy} measures the effect on the answers of the model.

\paragraph{Page plan.} A sliding-layer page of the 31B model holds 16 tokens $\times$ 16 heads $\times$ 256 dimensions $\times$ 2 (key and value) $\times$ 2 bytes $= 262{,}144$ bytes, and a stock global page holds 32 tokens $\times$ 4 heads $\times$ 1{,}024 values $\times$ 2 bytes, the same size. A compact global page of $B$ tokens needs $B \times 4 \times 640 \times 2$ bytes. The block size must be a multiple of 16 tokens, and the largest one that fits is $B = 48$ (245{,}760 bytes). The engine pads this page to 262{,}144 bytes, so 6.25\% of the padded page is unused, and the bytes per token of a global layer fall from 8{,}192 to 5{,}461, a factor of 1.5 against the ideal 1.6. The same plan gives a 48-token block for the 26B-A4B model and a 96-token block for the 12B model, with the same padding. An open vLLM proposal for a packed block stride with unequal page sizes~\cite{vllmrfc} could remove the padding if it is adopted.

\section{Implementation in vLLM}
\label{sec:impl}

We built the prototype on vLLM 0.33.0rc1.dev229+g3966924d5 with its Triton attention backend. The vLLM log states that it forces this backend for Gemma~4 on the A100, because the two layer types have different head sizes and FlashAttention-4 is not available there. The prototype has three parts, installed as patches before the engine starts:
\begin{enumerate}
\item \textbf{The KV specification.} The global layers report 640 values per head and token and a block of 48 tokens. The stock allocator, page unifier and scheduler then plan and manage the pool with no other change.
\item \textbf{The V view.} The Triton backend splits a cache row at the head size, so a 640-value row gives a correct K view and a value cache of only 128 values. Small wrappers of the attention and cache-write entry points replace that value cache with the V view (a strided view, no copy).
\item \textbf{The transforms.} Module hooks on the 10 global attention layers form the folded and permuted query of Eq.~\eqref{eq:fold} after RoPE, form the two write operands and permute the output back.
\end{enumerate}
The attention kernels are unchanged: we changed no kernel source code, and the wrappers and hooks change only Python calls. The run-time transforms launch about 110 small kernels per decode step of the 10 global layers. Because the weights are fixed, a production version could fold $w_\U$ into the query-norm weight and the output permutation into the input columns of $W_o$ at load time, which removes most of these kernels. These folds change the rounding or the order of the sums, so they need their own output check, which we did not run. The query permutation must follow RoPE, because RoPE pairs dimension $i$ with dimension $i + 256$. So either the permutation stays a run-time gather, or the RoPE operation takes a permuted index map.

\section{Evaluation}
\label{sec:eval}

\paragraph{Setup.} The engine runs of this section used one A100 SXM 80GB (Section~\ref{sec:node} uses two GPUs), vLLM 0.33.0rc1.dev229+g3966924d5, Triton 3.8.0 and the 16-bit \texttt{google/gemma-4-31B-it} checkpoint, with a 16-bit KV cache, no prefix caching and at most 2{,}048 batched tokens per step. The build, capacity and step-time runs used a fixed KV pool of 15~GiB (6{,}144 blocks) and at most 16 sequences; the 16-bit weights (62.6~GB) leave about this much of the GPU memory for the cache. The engine ran in eager mode unless we say otherwise. Appendix~\ref{app:repro} gives all settings.

\paragraph{Compared configurations.} We call each configuration that we compare an \emph{arm}. STOCK is the unmodified engine and cache. COMPACT is the prototype with the 640-value row. EMULV is the stock cache with the same query and output transforms as COMPACT: it computes the same arithmetic on the stock cache, so a difference between COMPACT and EMULV can come only from the cache layout. The accuracy study also has a noise arm, which adds one extra 16-bit rounding of the query to show how large a harmless rounding change looks, and a positive control, the stock 8-bit cache at a KV scale of 1.0, which we expected to change more answers.

\paragraph{Rules.} For each experiment we recorded the question and a pass-or-fail rule before the run, and we changed a rule only before the data of the run that it governs, with one disclosed exception (an FP8 test, Appendix~\ref{app:experiments}). We call a run \emph{invalid} if a validity check that we set before the run failed. This does not mean that its numbers are wrong, but we use them only as exploratory results.

\subsection{Exactness and accuracy}
\label{sec:accuracy}
COMPACT and EMULV gave bit-identical outputs for all 432 prompts below and for all 1{,}264 recorded arrays of prompt log-probabilities. So, on these prompts and in eager mode, the two-view read and the page plan add no error to the arithmetic of the transforms.

\paragraph{The prototype.} We measured the accuracy on 432 items: 400 GSM8K test problems~\cite{gsm8k} and 32 needle-retrieval prompts of 7.7K tokens, in chat format with greedy decoding. Every arm answered all 32 needle prompts correctly, so every change came from the GSM8K items. In the build run, COMPACT had an accuracy of 0.9769 and STOCK 0.9722 (422 and 420 of 432): a change of $+0.46$ points, with the 95\% Agresti--Min interval for paired answers $[-0.32, +1.24]$, inside our preset margin of $\pm 3$ points (we count a change inside this margin as no loss). Two final answers changed, both toward the correct answer, and the generated text differed in 88 of the 432 items (20.4\%). The mean negative log-likelihood that the model gives to the GSM8K reference solutions changed by $+0.00004$ nats per token (interval $[-0.0007, +0.0008]$). The stock engine alone scored 0.9769 in an earlier run with other settings (the batch cap, the KV pool and the maximum model length), so a change of 2 items has no stable sign.

\paragraph{An earlier emulation.} That earlier run emulated the compact arithmetic on the stock cache in the natural dimension order. It gave $-0.46$ points ($[-1.24, +0.32]$), with 2 changed answers against 3 for a noise arm that adds one extra 16-bit rounding of the query. Appendix~\ref{app:accuracy} gives its details, including a small shift of the reference log-likelihood whose sign we cannot explain.

\paragraph{Limits.} The result covers final answers, not text. The GSM8K sequences are short (about 240 tokens with the answer), and only the 32 needle prompts reach 7.7K tokens, so no accuracy test reaches the lengths at which the saving matters. The base accuracy is near the ceiling, and the positive control did not separate from the compact arm. So the result is ``no break found'', not a proof of equal quality.

\subsection{Memory and capacity}
\label{sec:capacity}

\begin{figure}[t]
\centering
\includegraphics[width=\linewidth]{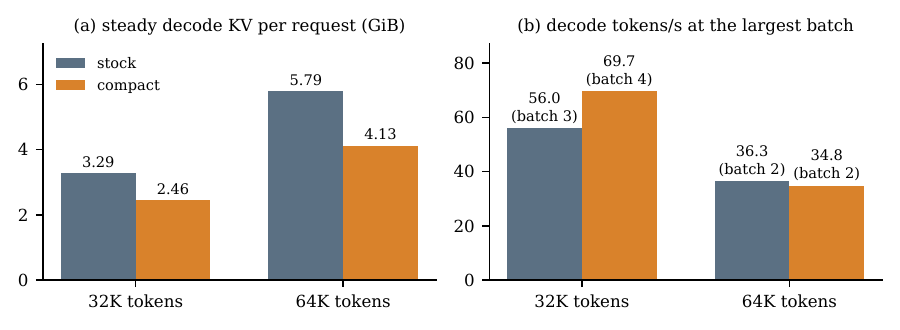}
\caption{Measured in vLLM on one A100 with the 16-bit 31B model and a 15~GiB KV pool, in eager mode. (a) The steady decode KV memory of one request, all layers. (b) The decode tokens per second at the largest batch that the engine admits and decodes with no preemption; the batch size is in parentheses.}
\label{fig:capacity}
\end{figure}

The \emph{steady decode footprint}, the steady decode KV memory of one request, is the number of blocks in use while that request decodes. The \emph{prefill peak} is the largest number of blocks in use while the engine prefills one request; it sets the cost of admission. The steady decode footprint fell from 1{,}349 to 1{,}008 blocks at 32K tokens and from 2{,}373 to 1{,}690 blocks at 64K tokens (Figure~\ref{fig:capacity}, Table~\ref{tab:capacity}): factors of 1.338 and 1.404, or 0.83 and 1.67~GiB less KV per request. These ratios count all layers and match the page-plan arithmetic to within one block: the sliding blocks do not change, and the global blocks fall by a factor of 1.5. So the saving is a factor of 1.6 in the global-layer values, 1.5 after the page padding, and 1.34 to 1.40 over all layers.

The saving at admission is smaller, because the prefill reservation includes up to 321 blocks in each sliding group in both layouts: the prefill peaks fell by factors of 1.15 and 1.23. At 32K tokens, the 15~GiB pool decoded 4 compact requests where 3 stock requests fit, and the decode rate rose from 56.0 to 69.7 tokens per second (1.24 times). At 64K tokens both layouts held 2 requests, and the compact build was 4\% slower; the third compact request lacked 181 blocks (about 0.44~GiB), so a slightly larger pool would probably hold 3. The gain is a decode-state saving that becomes throughput where one more request fits, and it depends on where the pool limit falls. By its preset rule (Appendix~\ref{app:experiments}), this run passed on outputs, accuracy and capacity, but not on throughput (a gain at one length only) or step cost (above the 5\% limit).

\begin{table}[t]
\centering
\small
\begin{tabular}{lcc}
\toprule
 & 32K tokens & 64K tokens \\
 & stock / compact & stock / compact \\
\midrule
Steady decode footprint (blocks) & 1{,}349 / 1{,}008 (1.338$\times$) & 2{,}373 / 1{,}690 (1.404$\times$) \\
Steady decode footprint (GiB, all layers) & 3.29 / 2.46 & 5.79 / 4.13 \\
Prefill peak (blocks) & 2{,}602 / 2{,}262 (1.150$\times$) & 3{,}626 / 2{,}944 (1.232$\times$) \\
Largest batch with no preemption & 3 / 4 & 2 / 2 \\
Decode tokens/s at that batch & 56.0 / 69.7 (1.24$\times$) & 36.3 / 34.8 (0.96$\times$) \\
Step-time ratio at the same batch (mean) & 1.087 & 1.043 \\
\bottomrule
\end{tabular}
\caption{Capacity in vLLM (one A100, 16-bit 31B, 15~GiB pool, eager mode). One block is one 262{,}144-byte page in each of the 10 layers of a group. The step-time ratio is compact over stock. Each throughput and step-time cell is one 128-step window of one engine process.}
\label{tab:capacity}
\end{table}

\subsection{Step time}
\label{sec:step}

In eager mode, the mean decode step of the prototype was 8.7\% slower than the stock step at the same batch at 32K tokens and 4.3\% slower at 64K. At 32K, the transforms cost 4.8\% (EMULV against STOCK), and the compact cache cost another 3.7\% (COMPACT against EMULV), measured in two engine processes. A benchmark of the global layers without weights found that the compact read was faster than the stock read, and that the transforms cost mostly CPU launch time; it explains about 40\% of the engine cost. The cause of the remaining cost is not known.

In an exploratory run with full decode CUDA graphs (the \texttt{FULL\_DECODE\_ONLY} mode of vLLM, without \texttt{torch.compile}), the compact step took 50.52~ms against 50.73~ms for the stock step at 3 requests of 32K tokens ($-0.41\%$, inside our 2\% noise band). A kernel profile agrees: the compact cache cut the attention kernel time by 0.49~ms per step, and the 110 transform kernels added 0.46~ms. The output check of this run failed for 1 of 3 prompts, probably because of a schedule difference outside the timing windows, so the run is invalid by its own rule. With full decode CUDA graphs we saw no step cost above about 2\%; we do not claim zero cost. Appendix~\ref{app:step} gives the details.

\subsection{A bound on the speed gain}
\label{sec:bytes}

A 37.5\% smaller global cache moves fewer bytes per decode step, so one could expect a faster step. At 3 requests of 32K tokens, one decode step reads about 62~GB (decimal gigabytes) of 16-bit weights and about 10.6~GB of KV (per request: 2.68~GB in the global layers and 0.84~GB in the sliding layers). The compact cache removes about 2.7~GB of these 73~GB, or 3.7\%. So even a step whose time followed its bytes would gain only about 4\% here. The profile of the exploratory graph run agrees: the matrix products took 37.4~ms and attention 10.7~ms of the 50.7~ms graph step, and the compact cache cut the attention time by 0.49~ms. At larger batches the KV share of the bytes grows, but in our kernel tests the attention time followed the DRAM bytes only weakly: on an A100 the compact read saved 2.7\% to 9.7\% of the read time, and on an H100 the FP8 kernels had a time elasticity of 0.16 to 0.24 to the bytes (exploratory; Appendix~\ref{app:kernels}). So the speed of one step does not change much; a gain in throughput must come from more requests in each step, which the next section measures.

\section{Capacity in a serving decode node}
\label{sec:node}

Section~\ref{sec:eval} measured fixed batches in one engine. Large deployments often split the work: a prefill engine computes the prompt, the KV cache moves to a decode engine, and the decode engine generates the output~\cite{distserve,splitwise,mooncake}. The decode engine of such a system holds many running requests, and its KV pool is full most of the time. When the pool is full, vLLM preempts a request; with a CPU offload connector, the request can later reload its blocks from host memory instead of computing them again. This section asks whether the compact cache works on this path, and whether its capacity gain matches a model of the block pool, with a decision rule that we froze before each run.

\subsection{A model of the block pool}
\label{sec:model}
We wrote a discrete-step model of the vLLM block pool for a decode engine. It follows the rules of the engine that we read in the source: the admission of a request only when its blocks for the full input fit, a token budget of 2{,}048 per step with two batches in flight (asynchronous scheduling), the preemption of the most recently admitted request when the pool is full, and a CPU tier that stores the blocks of all running requests (Appendix~\ref{app:node}). The input is a stream of requests with lognormal prompt and output lengths (log-scale standard deviation 0.6). The engine keeps a backlog of requests that wait for blocks, so the pool is the limit. A \emph{work step} is a step in which the scheduler schedules work; steps that only wait for a KV transfer do not count. The main statistic is
\begin{equation}
R_\text{step} = \frac{\text{decode tokens per work step, compact}}{\text{decode tokens per work step, stock}}\,,
\end{equation}
which counts how many more requests decode in each step and does not depend on the step time. The model gives $R_\text{step}$ for the exact request sequence that the engine runs (one random seed) and a range over seeds 0 to 4 and backlog depths 2, 3 and 8. As a report only, it also gives a throughput ratio with step times from attention read times that we measured on an A100 SXM (exploratory), for CUDA graphs and with the measured costs of eager mode added.

\subsection{The connectors}
\label{sec:connectors}
With the stock vLLM offload connector set to store all blocks, forced preemptions of compact requests on the 31B model reloaded their generated tokens from the CPU tier, and 24 checked reloads matched byte for byte, except blocks that vLLM had computed again. The NIXL connector~\cite{nixl} of the pinned vLLM build could not register the compact cache, and a byte view of each padded page removes the problem. By our reading of the code, vLLM main later fixed this check in general~\cite{vllmpr65685}, so our patch is likely only a back-port; we did not run vLLM main. With the 12B model, NIXL moved 8 long prompts to a decode engine with all blocks intact, and NIXL-fed compact requests also reloaded correctly. Appendix~\ref{app:node} gives the details.

\subsection{Setup}
\label{sec:nodesetup}
One container had two A100-SXM4-80GB GPUs: the prefill engine on one GPU and the decode engine, with NIXL and a 64~GiB CPU tier, on the other. Both engines ran the 16-bit 31B model in eager mode with no prefix caching; the decode engine used the 15~GiB pool of Section~\ref{sec:eval} and at most 16 sequences (Appendix~\ref{app:repro}). A driver kept two prefilled requests ready and three requests waiting on the decode engine, so the decode pool stayed full: the decode engine had a waiting request in at least 99.6\% of the work steps of every run. The prompts were random token ids, each request decoded its drawn output length (\texttt{ignore\_eos}), and decoding was greedy. Each arm measured the work steps 2{,}000 to 22{,}000. The stock arm ran the stock cache without the transforms; the compact arm ran the compact cache with them. We tested two operating points that differ in prompt length: D3H (median prompt 16{,}384 tokens) and D3Q (median prompt 8{,}192 tokens), both with a median output of 2{,}048 tokens. The model predicts a clearly smaller gain for the shorter prompts. We expected this: the sliding layers keep a fixed size, so the footprint ratio of Section~\ref{sec:capacity} falls with the prompt length.

\subsection{Results}
\label{sec:noderesults}

\begin{table}[t]
\centering
\small
\begin{tabular}{lccccc}
\toprule
 & \multicolumn{2}{c}{decode tokens per work step} & & \multicolumn{2}{c}{model prediction of $R_\text{step}$} \\
Run & stock & compact & $R_\text{step}$ & seed 0 & seeds 0--4 (frozen band) \\
\midrule
D3H, first run & 5.458, 5.459 & 7.509 & 1.376 & 1.364 & 1.308--1.364 (1.295--1.378) \\
D3H, repeat run & & 7.514 & 1.377 & & \\
D3Q & 8.192 & 10.133 & 1.237 & 1.229 & 1.169--1.229 (1.158--1.241) \\
\midrule
Model, seed 0 & 5.454 / 8.246 & 7.438 / 10.130 & & & \\
\bottomrule
\end{tabular}
\caption{The decode-node tests (two A100-SXM4-80GB, 31B, 15~GiB decode pool, eager mode). The first D3H run ran stock twice (before and after compact). The repeat run ran the compact arm again with a longer NIXL lease (no lease expired), and it is compared with the stock arms of the first run. The model row gives the running requests per step for D3H / D3Q. The frozen band is the seed range widened by at least 1\% on each side.}
\label{tab:node}
\end{table}

\begin{figure}[t]
\centering
\includegraphics[width=\linewidth]{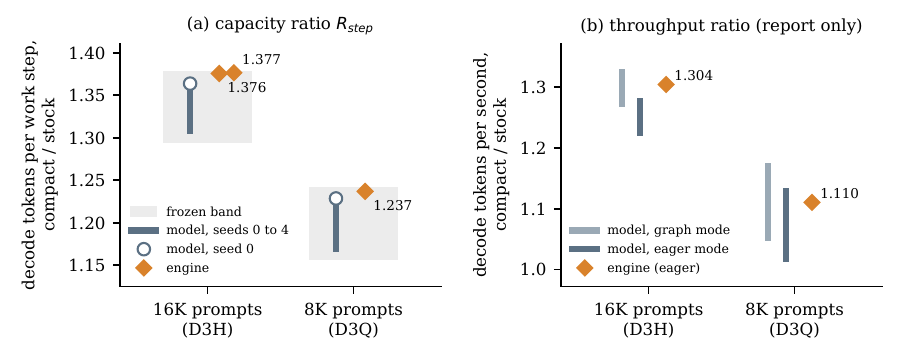}
\caption{The decode-node tests against the frozen model prediction. (a) The capacity ratio: the gray box is the frozen band, the bar the range of the model over 5 seeds and 3 backlog depths, the circle the model for the engine's request sequence, and the diamonds the engine runs. (b) The decode throughput ratio in eager mode against the model ranges with CUDA-graph and eager step costs (a report, single runs; the stock A/A spread of the throughput was 2.7\%).}
\label{fig:nodes}
\end{figure}

\paragraph{The capacity ratio.} Table~\ref{tab:node} and Figure~\ref{fig:nodes} give the results. At D3H the compact decode engine ran 1.376 times the stock decode tokens per work step; the model predicted 1.364 for the same request sequence. The two stock runs agreed to 0.03\%, and a second compact run (the \emph{repeat run}), in another session on another host, repeated the compact value to 0.06\%. At D3Q the ratio was 1.237, against 1.229 in the model. Both values are inside their frozen bands, but only because the bands were widened by 1\%. The engine was 0.9\% (D3H) and 0.7\% (D3Q) above the top of the model range. The engine ran the request sequence of seed 0, which is at the top of that range at both points. Other seeds give ratios up to 4\% to 5\% lower, so the frozen rule would also have passed a model that overstated the gain by that much.

The ratio fell by 0.139 from D3H to D3Q, and the model predicted a fall of 0.135. So the measured fall was 1.03 times the predicted one, inside the frozen limits of 0.75 to 1.25. This result does not depend on which D3H run we use: the limits hold for any D3H value from 1.338 to 1.406. The test confirms the direction and the size of the fall, not an accuracy of 1\%.

The model does not predict each arm exactly: at D3H the compact engine was 0.95\% above it, and at D3Q the stock engine was 0.6\% below it, while the other arm matched (Table~\ref{tab:node}). So the error of the model is not a constant bias of one layout (Appendix~\ref{app:node}).

\paragraph{Throughput.} The eager decode throughput rose by a factor of 1.30 at D3H and 1.11 at D3Q. In these runs a compact step took 5.5\% (D3H) and 11\% (D3Q) longer than a stock step, but it also held more requests; at equal batch the cost was 4\% to 9\% (Section~\ref{sec:step}). Both values lie inside the range that the model gives with CUDA-graph step costs, and the D3Q value also inside the eager range. These are single runs: the two stock runs at D3H differed by 2.7\% in throughput, so we do not resolve differences smaller than about 3\%. Compared with Section~\ref{sec:capacity}, where a fixed batch gained only where one more request fitted, a stream of requests of mixed lengths turns the smaller footprint into a steady gain.

\paragraph{Exact output on this path.} In the session of the repeat run, we also compared the compact engine with EMULV (Section~\ref{sec:eval}) on the same two-GPU path: the prefill engine prefilled 8 needle prompts of 7{,}666 tokens one at a time, NIXL moved their KV caches, and the decode engine decoded each alone, 64 tokens with \texttt{ignore\_eos}. The 512 output tokens were identical, the log-probabilities of the sampled tokens were bit-equal, and both engines gave the correct answer for 8 of 8 prompts. Only the first 4 to 6 tokens of each output are the answer; the rest are forced text that still tests the numerical identity.

\paragraph{A harness defect in the first D3H run.} In the first D3H run, the NIXL lease of 35\% to 42\% of the requests expired before the decode engine read them, so these requests may have decoded with other KV values. This cannot change the count of decode tokens per step, and the repeat run, with a longer lease and no expired lease, repeated the compact value to 0.06\% (Appendix~\ref{app:node}).

\section{Related work}
\label{sec:related}

\paragraph{Smaller KV caches by architecture.} Multi-query~\cite{mqa} and grouped-query attention~\cite{gqa} share keys and values across query heads, and cross-layer schemes share them across layers~\cite{cla,yoco}. Multi-head latent attention~\cite{deepseekv2} caches a compressed latent and a separate RoPE key, and its decode kernels~\cite{flashmla} read the key and the value from one cached row. Grouped-tied attention~\cite{gta} and grouped value attention~\cite{gva} also keep one cached state with a separate RoPE part, but both are trained from scratch; MHA2MLA~\cite{mha2mla} and TransMLA~\cite{transmla} convert trained models to MLA with fine-tuning. The Gemma~4 report credits Kayyam et al.~\cite{kayyam} for the reuse of keys as values in global layers. Our row has the shape of these designs, but it is an inference layout for released models that already tie the key to the value, with no training and no new kernel. The fold of $w_\U$ into the query is the absorption technique of MLA~\cite{deepseekv2}.

\paragraph{KV compression at inference.} Quantization~\cite{kivi,kvquant} and token eviction~\cite{h2o,snapkv,streamingllm} reduce the cache with some loss. LowRAM~\cite{lowram} predicts the values from the keys and stores residuals, which is near-lossless. Slim attention~\cite{slimattention} stores only the keys and computes the values from them; it is exact in real arithmetic, but it needs an invertible key projection and extra compute. The Gemma~4 global layers are an elementwise case of such a relation, so the compact row needs no rebuild of the values. Kanwat~\cite{foldbias} also folds a term into the keys so that unchanged vLLM kernels read a compacted cache.

\paragraph{Serving engines.} PagedAttention and vLLM~\cite{vllm} manage the cache in blocks, and Jenga~\cite{jenga} manages layers with different page sizes in vLLM; our page plan is a small case of that problem. Disaggregated serving moves the KV cache from prefill GPUs to decode GPUs~\cite{distserve,splitwise}, and Mooncake~\cite{mooncake} also keeps KV caches in CPU memory and SSDs. These systems change where the cache lives and when it moves; the compact row makes the cache of each request smaller, so it combines with them (Section~\ref{sec:node}). We read the KV-cache code of vLLM, SGLang~\cite{sglang}, TensorRT-LLM, Hugging Face transformers~\cite{transformers} and other code bases, and we searched 15 repositories, arXiv and the web (Appendix~\ref{app:repro}). In each code base, these layers write a full key and a full value, or the model conversion produces standard key and value heads. We found no code, issue or pull request that stores the Gemma~4 global KV cache once; a search cannot see private code.

\section{Limitations}
\label{sec:limits}
\begin{itemize}
\item One model family and one vLLM build. All measurements use the 31B model, except the NIXL connector tests of Section~\ref{sec:connectors}, which used the 12B model; the 26B-A4B model has none. The method needs this K=V structure, and such exact redundancy is rare. In 16 current model configurations that we read on 2026-11-10, every other exact redundancy of 10\% or more that a model declares in its KV cache was already used by vLLM or by the serving image that the model supports.
\item The decode-node tests (Section~\ref{sec:node}) cover two operating points on one axis (the prompt length) at one pool size, one request sequence per point, eager mode, one cloud host per run and random-token prompts with fixed output lengths. Both point checks pass only because the frozen bands were widened by 1\%: the engine was 0.7\% to 0.9\% above the range of the model. The error of the model differs by point and by layout, so we do not claim a constant bias. The throughput ratios are single runs, with a stock A/A spread of 2.7\%. The NIXL patch is likely only a back-port for the pinned vLLM build, and the exactness check on this path used 8 prompts of one length at batch 1.
\item The engine tests used a 16-bit cache and only the Triton backend on an A100. The FP8 compact row and the FlashAttention-4 read have only kernel tests on an H100 and a CPU test on stored activations, and at the default FP8 KV scale of 1.0 the compact row fails our attention-error bound, so the saving by a factor of 1.34 to 1.40 applies only to a 16-bit cache. With one FP8 scale per row, the compact row had 1.06 times the pooled attention error of the per-token-head FP8 cache of vLLM (up to 1.18 times in one layer), with no engine and no model-level test (Appendix~\ref{app:kernels}).
\item Each test of Section~\ref{sec:eval} used one GPU, and the A100 ran under its power cap. Its capacity and timing runs used synthetic random-token prompts of equal length. We did not test prefix caching (the 48-token block changes the alignment of cache hits), tensor parallelism, speculative decoding or more than 16 concurrent sequences; Section~\ref{sec:node} used mixed-length traffic, but only on one decode node.
\item The accuracy test covers final answers with greedy decoding; its longest prompt has 7.7K tokens, its margin of $\pm 3$ points is a project heuristic, and its positive control did not separate from the compact arm. In our vLLM path, the instruct model scored raw natural text poorly in both checkpoints that we used (up to 10.5 nats per token on books); we did not find the cause or run a reference check with Hugging Face transformers, and all our model-level results use this path. Because both arms use the same path, a shared defect would not create a difference between them, but it could hide one.
\item For a fixed batch, the capacity gain is an integer effect of the pool size: at 64K tokens it gave no throughput gain in our pool. A stream of requests of mixed lengths gave a steady gain (Section~\ref{sec:node}).
\item The prototype applies its transforms at run time. The load-time folds and the default graph mode of vLLM (piecewise graphs with \texttt{torch.compile}) were not tested, the CUDA-graph timing is exploratory, and we had no hardware counters for DRAM traffic.
\end{itemize}

\section{Conclusion}
The global KV cache of Gemma~4 can be stored once, exactly in real arithmetic, as one 640-value row per KV head and token that unchanged attention kernels read through two views. In vLLM this cuts the steady decode KV memory per request by a factor of 1.34 to 1.40 at 32K to 64K tokens, and the final answers did not change beyond noise in our tests at contexts up to 7.7K tokens. The method is a memory mechanism: it adds requests per GPU where the KV pool is the limit. On a prefill/decode pair with NIXL transfer and CPU offload, it gave a decode engine with a full pool 1.38 and 1.24 times the decode tokens per step at two operating points, as a model of the block pool predicted, and 1.30 and 1.11 times the eager decode throughput. On this path its output matched its stock emulation for 8 prompts at batch 1. The run-time prototype costs 4\% to 9\% of an eager step at equal batch, and in an exploratory run with full decode CUDA graphs we saw no step cost above about 2\%. We hope that serving engines adopt the layout for Gemma~4 and for future models that tie keys to values.

\paragraph{Code and data.} The prototype code and the experiment records are at \url{https://github.com/Kushagrabainsla/gemma4-compact-kv}.

\paragraph{Acknowledgments.} We thank Rishabh Sinha for suggesting one FP8 scale per compact row (Appendix~\ref{app:kernels}).

\bibliographystyle{unsrtnat}
\bibliography{refs}

\appendix
\section{The experiments}
\label{app:experiments}
Each experiment had a written question and a decision rule that we froze in the project repository before the run. The design was reviewed before the run, and amendments came before the data of the run that they govern (one exception: E-74, below). For the GPU runs, a rehearsal with a smaller model or on a cheaper GPU ran first, and a review after the run recomputed the results. E-70 itself used the 12B model; E-72 and E-73 had no separate rehearsal, because they reused the code of E-71. Table~\ref{tab:experiments} lists the experiments with the outcome that the frozen rule gave. The outcome words are those of each frozen rule. GO and PASS mean that the conditions of the rule held. PIVOT means that they held only in part, so the next step changed. BUILD-PIVOT (E-57) means that the outputs, the accuracy and the capacity passed, but the throughput gain held at one length only and the step cost was above its 5\% limit. FAIL means that the result missed its bound. INVALID means that a validity check of the design failed; this does not mean that the data are wrong. We use numbers from INVALID runs only as exploratory results. The words of the decode-node tests (Section~\ref{sec:node}): CONNECTOR-WORKS, PATCH-WORKS and RELOAD-WORKS mean that the transfer and reload checks passed, and GAP-CONFIRMED that the unpatched connector failed as predicted; STEP-CONFIRMS means that $R_\text{step}$ was inside the frozen band (CLEAN: with no expired NIXL lease); REPEATS means that a second run was within 0.5\% of the first; SHIFT-MATCHES means that the change of $R_\text{step}$ between two operating points was 0.75 to 1.25 times the predicted change; EXACT means identical output tokens. E-63, E-64 and E-67 are models that ran on a CPU. E-74, run after the first version of this paper, is a CPU test on the stored activations of E-54. Its frozen rule first gave INVALID: one validity check used a relative limit, which cannot hold in the layer where every value of $w_\U$ is $2^{-4}$, because there the reference error is float32 noise. After the run we added an absolute floor of $10^{-6}$ to that check, computed nothing again, and the rule gave GO with the same decision statistics. The main text names the decode-node runs by their role: the first D3H run is E-71, the repeat run is E-72, and the D3Q run is E-73. The main text writes ``invalid'' for the outcome INVALID.

In E-59 the output check required identical tokens from COMPACT and EMULV for 3 prompts. For one prompt the tokens differed from position 3, both with CUDA graphs and in eager mode. The compact cache leaves more blocks free, so the compact engine admitted that request one step earlier than the engine with the stock cache, and its prompt was split into other prefill chunks. The prompt is a random-token prompt, and STOCK and EMULV also differ on it (from position 4). A review of the run found the schedule difference sufficient to explain the failure; we did not show that it is the only cause. The two stock processes of E-59 differed by 1.1\% in eager windows that we chose after the run; its frozen analysis, which included windows after a profiler pass, recorded 6.7\%.

\begin{table}[p]
\centering
\footnotesize
\begin{tabular}{ll>{\raggedright\arraybackslash\hyphenpenalty=10000}p{3.4cm}>{\raggedright\arraybackslash}p{2.7cm}>{\raggedright\arraybackslash}p{5.0cm}}
\toprule
Run & GPU & Outcome & Used in this paper as & What it measured \\
\midrule
E-49 & H100 & PIVOT & context (App.~\ref{app:kernels}) & Ceiling of the decode-step saving, 16-bit KV (6.65\% and 11.33\%) \\
E-50 & H100 & GO & context (App.~\ref{app:kernels}) & The same on an FP8 KV cache near the pool limit (12.1\% and 15.0\%) \\
E-51 & H100 & INVALID & exploratory (App.~\ref{app:kernels}) & A compact FP8 Triton kernel against its standard twin \\
E-52 & H100 & INVALID & exploratory (App.~\ref{app:kernels}) & The two-view read in Triton and stock FP8 kernels \\
E-53 & H100 & FAIL & result (App.~\ref{app:kernels}) & The FP8 compact row at KV scale 1.0, real activations \\
E-54 & H100 & PASS & result (App.~\ref{app:kernels}) & The FP8 compact row with tied, calibrated scales \\
E-55 & H100 & INVALID & not used & A first model-level test (the baseline could not resolve the margin) \\
E-56 & A100 & GO & result (Sec.~\ref{sec:accuracy}, App.~\ref{app:accuracy}) & Accuracy of the emulated compact cache \\
E-57 & A100 & BUILD-PIVOT & main result (Sec.~\ref{sec:accuracy} to~\ref{sec:step}) & The vLLM prototype: exactness, accuracy, capacity, eager step time \\
E-58 & A100 & GO & result (Sec.~\ref{sec:step}, App.~\ref{app:step}) & Benchmark of the global layers without weights \\
E-59 & A100 & INVALID & exploratory (Sec.~\ref{sec:step}, App.~\ref{app:step}) & The prototype with full decode CUDA graphs \\
E-63 & CPU & PIVOT-V2 & context (Sec.~\ref{sec:model}) & The block-pool model: decoding requests per step \\
E-64 & CPU & no GO & context (Sec.~\ref{sec:model}) & The model on request streams, unified and decode nodes \\
E-67 & CPU & OFFLOAD-CONFIG-GO & context (Sec.~\ref{sec:model}) & The model with the CPU offload connector \\
E-68 & A100 & INVALID & exploratory (Sec.~\ref{sec:model}) & Attention read times for the model's step times \\
E-69 & A100 & CONNECTOR-WORKS & result (App.~\ref{app:node}) & The offload connector with the compact cache \\
E-70 & 2$\times$L40S & PATCH-WORKS, GAP-CONFIRMED; RELOAD-WORKS & result (App.~\ref{app:node}) & NIXL transfer of the compact cache (12B); reload of NIXL-fed requests \\
E-71 & 2$\times$A100 & STEP-CONFIRMS & main result (Sec.~\ref{sec:node}) & The decode node at D3H \\
E-72 & 2$\times$A100 & REPEATS, STEP-CONFIRMS-CLEAN, EXACT & result (Sec.~\ref{sec:node}) & The repeat with a long lease; exact output on the path \\
E-73 & 2$\times$A100 & STEP-CONFIRMS, SHIFT-MATCHES & main result (Sec.~\ref{sec:node}) & The decode node at D3Q \\
E-74 & CPU & INVALID, then GO after a post-run change of one check & result (App.~\ref{app:kernels}) & One FP8 scale per compact row, on stored activations \\
\bottomrule
\end{tabular}
\caption{The experiments behind this paper and the outcomes of their frozen rules. The column ``Used in this paper as'' says how the paper uses each run: as the main result, as a result, as an exploratory result (the INVALID runs), as context or not at all. The GPU time of E-49 to E-59, with their rehearsals, cost about \$21, and that of E-68 to E-73 about \$28.}
\label{tab:experiments}
\end{table}

\section{Accuracy details}
\label{app:accuracy}
The earlier accuracy run (E-56) emulated the compact arithmetic on the stock cache in the natural dimension order, so its sums run in a different order from the build. It used an automatic pool (92\% of the GPU memory), at most 64 sequences and a maximum model length of 10{,}240 tokens; the build run used the 15~GiB pool, at most 16 sequences and 66{,}048 tokens. It had a noise arm, which adds one extra 16-bit rounding of the query, and a positive control, the stock 8-bit cache at a KV scale of 1.0. The accuracy change was $-0.46$ points ($[-1.24, +0.32]$). Two of 432 final answers changed, against 3 for the noise arm; one item loops in the stock run and counts once in each compared arm, and without it the counts are 1 and 2. The mean negative log-likelihood of the reference solutions changed by $-0.0009$ nats per token (interval $[-0.0016, -0.0002]$); the noise arm moved it by $+0.0027$, outside our margin of 0.002, and the sign of the compact shift is unexplained. The generated text differed in 21.5\% of the items (noise arm: 17.4\%). Over the needle-prompt tokens after position 1{,}024, the mean negative log-likelihood rose by $+0.0066$ nats per token (interval $[+0.0005, +0.0127]$) under the emulated compact cache, by $+0.0074$ under the noise arm and by $+0.0066$ under the 8-bit control. But the baseline there is 11.04 nats per token, the poorly scored raw-text regime of Section~\ref{sec:limits}, so these numbers support no conclusion about the accuracy at long context.

\section{Step-time details}
\label{app:step}
\paragraph{Eager mode.} The median slowdowns of the eager build run are 7.9\% (32K) and 2.1\% (64K): the step times are bimodal (20\% to 32\% of the steps are about 5~ms slower than the median), and the compact arm has a slightly heavier slow tail.

\paragraph{The attention layer.} A benchmark of the 10 global layers with synthetic caches and no model weights timed each part on its own, at the shapes of the engine runs. Read by the stock Triton kernel, the compact cache as built was 0.49~ms (32K) and 0.23~ms (64K) faster per 10-layer step than the stock cache, so the compact read does not explain the 3.7\%. The transforms took 3.14~ms of CPU launch time per 10-layer step, but only 0.40~ms of GPU time inside a CUDA graph. In wall time they added 1.04~ms per 10-layer step, against 2.55~ms in the engine, so the benchmark explains about 40\% of the engine cost. This agrees with a launch-time cost, but we did not test whether the eager engine step is bound by the CPU.

\begin{figure}[t]
\centering
\includegraphics[width=0.72\linewidth]{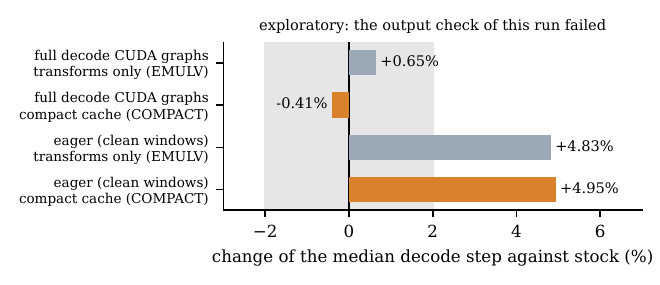}
\caption{Change of the median decode step time against the stock engine in the same mode, at 3 requests of 32K tokens (31B, one A100). The gray band is $\pm 2\%$. Exploratory: the output check of this run failed, for a reason that we think lies outside the timing windows. We chose the eager 128-step windows after the run, from windows that the profiler did not touch; the compact and stock eager windows come from different engine processes (two stock processes differed by 1.1\%), and within one process the spread was 0.09\%.}
\label{fig:step}
\end{figure}

\paragraph{The engine with CUDA graphs.} We then ran the 31B engine with full decode CUDA graphs (the \texttt{FULL\_DECODE\_ONLY} mode of vLLM, without \texttt{torch.compile}), with the transforms attached before the graph capture (Figure~\ref{fig:step}). At 3 requests of 32K tokens, the compact step took 50.52~ms against 50.73~ms for the stock step (medians of 128-step windows; $-0.41\%$), and the transforms alone added 0.65\%. The noise was 0.09\% within a process and 0.13\% between processes, so $-0.41\%$ is inside our 2\% band and is not a speedup. A kernel profile agrees: the compact cache cut the attention kernel time by 0.49~ms per step, and the 110 transform kernels added 0.46~ms. The A100 lowers its clock when it reaches its power cap; the compact engines drew less power and ran at a 15~MHz higher clock. We count this as part of the effect of the method, not as a kernel gain. In the eager 128-step windows of the same run that the profiler did not touch (we chose them after the run), the transforms cost 4.8\% again and the whole compact step 4.95\%, against 8.7\% in the first measurement. The cache-path cost did not appear there ($+0.1\%$; two stock processes differed by 1.1\%), so the cause of the 3.7\% is not known.

We label this run \emph{exploratory}: its frozen output check failed for 1 of 3 prompts. A review of the run found that the compact engine, with more free blocks, admitted that request one step earlier and split its prompt into other prefill chunks; this is sufficient to explain the failure, but we did not show that it is the only cause. The check assumed equal schedules, so the run is INVALID by its own rule. The timing windows contain only full decode steps of all 3 requests, in every arm (Appendix~\ref{app:experiments}). With full decode CUDA graphs we saw no step cost above about 2\%; we do not claim zero cost.

\section{Kernel and FP8 studies}
\label{app:kernels}
\paragraph{Ceilings.} The H100 studies used one H100 80GB and the FP8 checkpoint of Appendix~\ref{app:repro}, a different operating point from the A100 engine runs. The ceiling estimate of the decode-step saving (at the lower 95\% bound of the time slope) was 6.65\% at 32K and 11.33\% at 64K tokens with a 16-bit KV cache at batch 4 (E-49), and 12.1\% and 15.0\% with an FP8 KV cache near the pool limit (E-50). These estimates assume that the attention time follows the bytes.

\paragraph{Kernel time.} The attention time does not follow the bytes: in three families of FP8 attention kernels for head size 512, the time elasticity to DRAM bytes was 0.16 to 0.24, so removing a fraction $f$ of the bytes removed only about $0.2f$ of the time that grows with the context. The stock FlashAttention-4~\cite{flashattention} FP8 kernel of vLLM read the two views of an FP8 compact cache with no code change and with the same error as its normal input. It was 7\% to 10\% faster in the context-dependent time, which is about 3\% of a decode step near the pool limit, and its measurements also ran at a 5.7\% higher clock. These two kernel runs (E-51 and E-52) are INVALID by their frozen rules, so their numbers are exploratory. On the A100 (E-58), the compact row read in a 32-token block saved 6.6\% (32K) and 3.5\% (64K) of the read time, and the 48-token cache as built saved 9.7\% and 2.7\%. In rehearsal runs on GPUs with less bandwidth per unit of compute, the saving was larger: 28\% to 37\% of the read time on an A40 (the same benchmark) and 35\% to 36\% of the context-dependent kernel time on an RTX 4090. This suggests that the speed effect grows on such GPUs, but each value comes from one rehearsal, and these GPUs also differ in other ways.

\paragraph{An FP8 compact row.} We tested the compact row with an FP8 cache on real activations of the global layers, captured in vLLM on one H100. At the default KV scale of 1.0, the compact row failed our attention-error bound in one layer. With tied, calibrated per-layer scales, it had 0.67 times the attention error of the stock FP8 cache and 1.13 times the error of the best-calibrated standard FP8 cache, on 7 prompts. It needs scales with an error of at most 0.1\% to 0.25\%, and the stock FlashAttention-4 FP8 kernel read it in decode calls. The tied row has no model-level test, so we do not know whether it stacks with FP8 quantization.

\paragraph{One FP8 scale per row.} After the first version of this paper, we tested one FP8 scale per row (E-74): each 640-value row takes the scale $\max|r|/448$, the dynamic per-token-head rule of vLLM, so the two views share one scale with no calibration. We computed the attention in float32 on a CPU from the stored activations of the 7 prompts above (codes times scale). At the exact rule, this row had 1.06 times the attention error of the standard cache with per-token-head scales, and an emulation of the vLLM Triton read gave the same ratio. Over eight common scale factors $2^{j/8}$ ($j = 0, \ldots, 7$; $j = 0$ is the exact rule) applied to both caches, the ratio had a median of 1.02 and a range of 0.98 to 1.06. No row had its largest value in $k_\R$, and 0.4\% of the $k_\R$ codes were subnormal, so the rotary part does not make one scale too coarse. At the exact rule, the compact row with one scale per row had 0.46 times the error of the stock FP8 cache (scale 1.0) and 0.69 times the error of the calibrated tied row; the per-token-head cache had 0.43 and 0.65 times. This advantage holds only at the exact rule: a scale 9\% above the rule ($j = 1$) doubled the error of both caches, and over the eight factors the median error of the compact row was about that of the stock cache (0.99 times). An exploratory check on 5 of the 70 layer and prompt cells, for the per-token-head cache only, suggests that in 4 of them the exact storage of the largest value of each row explains most of the error increase at a shifted scale. In an engine, the two views would need to share one scale slot per row, which we did not build, and this test has no model-level part.

\section{Decode-node details}
\label{app:node}
\paragraph{CPU offload (E-69).} We used the stock vLLM offload connector set to store all blocks (\texttt{offload\_\allowbreak prompt\_\allowbreak only} off and \texttt{prefix\_cache\_\allowbreak retention\_\allowbreak interval} unset; with the default retention of 0, a reload stopped at the end of the prompt in a 12B control). We forced preemptions of compact requests on the 31B model on one A100. The reloads restored the generated tokens (a median decode coverage of 0.977; stock 0.984), and 24 checked reloads (11{,}676 blocks) matched byte for byte, except blocks that vLLM had computed again after an earlier reload: the tier keeps the first stored copy of a block and does not store it again, so these came back as the first version. Loads ran at about 24~GB/s.

\paragraph{NIXL transfer (E-70).} The padded compact page does not divide the shared KV buffer into NIXL regions, so the connector of the pinned build refused the cache. Our patch registers a byte view of each padded page (a two-dimensional \texttt{uint8} view at the same address). With the 12B model on two L40S, NIXL moved 8 prompts of 7{,}666 tokens from a prefill engine to a decode engine; all 3{,}192 full blocks matched, and the decode engine answered 8 of 8 correctly. With both connectors on one decode engine, forced preemptions reloaded NIXL-fed compact requests with their generated tokens (a median coverage of 0.953, 24 checked reloads of 9{,}192 blocks, with no mismatch other than the first-version effect).

\paragraph{The CPU tier of the model.} When the tier is full, the model evicts the request with the oldest rank, which is set at its first store and renewed when the request finishes. The model tier holds 200~GB and the engine tier 64~GiB; a check with a 64~GiB and a 10~GB tier gave the same predictions.

\paragraph{The model error for each arm.} At D3H the stock engine matched the model (5.458 against 5.454 decode tokens per work step), and the compact engine was 0.95\% above it. At D3Q the compact engine matched (10.133 against 10.130), and the stock engine was 0.6\% below. The engine also preempted more than the model (6 against 3 per arm at D3H; 12 and 16 against 8 and 9 at D3Q); we did not test whether this explains the difference.

\paragraph{The NIXL lease in E-71.} In E-71 the driver held prefilled requests outside the decode scheduler for up to a few minutes. The prefill engine keeps the blocks of a prefilled request for a lease of 30~s by default, and the decode engine extends the lease only for requests in its own scheduler. So 35\% to 42\% of the requests were read after the prefill engine had freed their blocks. The count of decode tokens per step does not depend on the KV values (the output lengths are fixed and prefix caching is off), and both layouts were affected alike. E-72 used a 600~s lease, and no lease expired. In E-72 the decode-engine logs also showed that the 8 exactness prompts came through NIXL and were not computed again.

\section{Model dimensions and reproducibility}
\label{app:repro}
\paragraph{Dimensions of Gemma~4 31B.} Hidden size 5{,}376; 60 layers (50 sliding-window layers with window 1{,}024, head size 256 and 16 KV heads; 10 global layers with head size 512 and 4 KV heads); 32 query heads; RoPE on the global layers with $p = 0.25$ and base $10^6$. At 32K tokens, the global KV of one request takes 10 layers $\times$ 32{,}768 tokens $\times$ 8{,}192 bytes $= 2.50$~GiB in the stock cache, and 1.67~GiB in the compact cache with the 48-token page plan.

\paragraph{Software and data versions.} The runs used these versions.
\begin{center}\scriptsize
\begin{tabular}{@{}ll@{}}
vLLM, Triton & 0.33.0rc1.dev229+g3966924d5; 3.8.0\\
Image & \texttt{vllm/vllm-openai@sha256:f255c4fad2f30ff5ee80db14d3a03afe9ae0f10d3470339864ff04745fe44cb3}\\
Checkpoint & \texttt{google/gemma-4-31B-it}, revision \texttt{842da3794eaa0b77d5f08bae87a17459d91ff475}\\
H100 checkpoint & \texttt{RedHatAI/gemma-4-31B-it-FP8-dynamic} (FP8)\\
GSM8K & \texttt{openai/gsm8k}, revision \texttt{740312add88f781978c0658806c59bc2815b9866}\\
12B checkpoint & \texttt{google/gemma-4-12B-it}, revision \texttt{43c9d014b231474fa7212dbb48a762566c2b255d}\\
NIXL & 1.5.0 (in the same image)\\
\end{tabular}
\end{center}

\paragraph{Engine settings.} One A100 SXM 80GB, eager mode unless stated, a 16-bit KV cache, no prefix caching, seed 0 and at most 2{,}048 batched tokens per step. The build, capacity and step-time runs (E-57 to E-59): a fixed KV pool of 15~GiB (6{,}144 blocks), at most 16 sequences and a maximum model length of 66{,}048 tokens. The emulation run (E-56): an automatic pool (92\% of the GPU memory), at most 64 sequences and a maximum model length of 10{,}240 tokens.

\paragraph{Decode-node settings (E-69 to E-73).} Modal containers (gVisor) with one A100-SXM4-80GB (E-69), two L40S (E-70) or two A100-SXM4-80GB (E-71 to E-73); the prefill and decode engines run in two processes on separate GPUs of one container. NIXL uses UCX with \texttt{ucx\_error\_handling\_mode} set to \texttt{none} (the container has no TCP network devices, and the default mode refused the intra-node transports). The decode engine uses a MultiConnector of the NIXL connector and the offload connector (\texttt{cpu\_bytes\_to\_use} 64~GiB in E-71 to E-73, \texttt{offload\_prompt\_only} off, \texttt{prefix\_cache\_retention\_interval} unset). E-71 to E-73: decode pool 15~GiB, at most 16 sequences and 2{,}048 batched tokens; prefill engine at most 4 sequences and 1{,}024 batched tokens; maximum model length 144{,}000 tokens on both; NIXL lease 30~s (the default) in E-71 and 600~s in E-72 and E-73. The request streams: prompt lengths lognormal with median 16{,}384 (D3H) or 8{,}192 (D3Q) tokens, clipped to 256 to 131{,}072; output lengths lognormal with median 2{,}048, clipped to 16 to 16{,}384; log-scale standard deviation 0.6; random number seed string \texttt{64-D3H-0} or \texttt{64-D3Q-0}; requests whose stock reservation does not fit the empty pool, or whose prompt and output exceed the maximum model length, are skipped, in the model and in the engine alike.

\paragraph{Accuracy data.} We drew 400 problems from the GSM8K test split with seed 56. The user turn is ``Solve the math problem. Show brief steps, then write the final answer as a number after `\#\#\#\#'.'' and the problem. The 32 needle prompts have 7{,}666 to 7{,}668 tokens: 7{,}600 tokens from one of 4 PG-19 test books, with the sentence ``The secret passphrase for the archive is <word>-<4 digits>.'' at 8 depths (5\% to 89\%), and the question ``What is the secret passphrase for the archive? Answer with the passphrase only.'' The prompts use the chat template of the checkpoint with thinking off. Decoding is greedy, with at most 640 tokens for GSM8K and 128 for the needles, and it stops at the end-of-turn token (id 106).

\paragraph{Patch points.} (1) The KV-cache specification of each global layer: 640 values (1{,}280 bytes) per head and token, and a block of 48 tokens. (2) Wrappers of the Triton attention entry point and of the Triton cache-write entry point, which replace that 128-value cache with the V view. (3) Hooks on the 10 global attention modules. They form the query $[\,q_\R \mid q_\U \odot w_\U\,]$ after RoPE (the product is rounded to 16 bits), the write operands $[\,k_\R \mid v_\U\,]$ and $[\,v_\U \mid v_\R\,]$, and the output permutation back to the model order. In the row, $\R$ is dimensions 0 to 63 and 256 to 319 in this order, and $\U$ is dimensions 64 to 255 and 320 to 511. For weights that are quantized in blocks of columns, an output fold must keep the blocks intact; the segments of the permutation start at multiples of 64 dimensions.

\paragraph{Search record.} On 2026-11-17 we repeated web searches for the method (the web, arXiv, and the vLLM and SGLang issues and pull requests) and found no new work that stores the Gemma~4 global KV cache once. Between 2026-10-25 and 2026-11-05, we read code in, or searched the issues and pull requests of, these 15 GitHub repositories, with fixed queries such as \texttt{attention\_k\_eq\_v}, \texttt{k\_eq\_v}, \texttt{gemma4 kv cache}, \texttt{gemma4 compact kv}, \texttt{head\_size\_v} and \texttt{kv cache keys values once} (buun-llama-cpp is a llama.cpp fork with a coupled key-value cache for latent attention):
\begin{center}\footnotesize
\begin{tabular}{@{}lll@{}}
\texttt{vllm-project/vllm} & \texttt{vllm-project/tpu-inference} & \texttt{vllm-project/flash-attention}\\
\texttt{sgl-project/sglang} & \texttt{NVIDIA/TensorRT-LLM} & \texttt{ggml-org/llama.cpp}\\
\texttt{spiritbuun/buun-llama-cpp} & \texttt{huggingface/transformers} & \texttt{google/gemma.cpp}\\
\texttt{google-deepmind/gemma} & \texttt{google-ai-edge/LiteRT-LM} & \texttt{google-ai-edge/litert-torch}\\
\texttt{flashinfer-ai/flashinfer} & \texttt{LMCache/LMCache} & \texttt{ml-explore/mlx-lm}\\
\end{tabular}
\end{center}
Between 2026-10-28 and 2026-11-05, we read the attention and KV-cache code of vLLM, its TPU backend, SGLang, TensorRT-LLM, Hugging Face transformers and gemma.cpp, and we checked llama.cpp, the buun-llama-cpp fork, Google DeepMind's JAX Gemma library and LiteRT-Torch. We found work that removes the duplicated value weights and work that adds FP8 caches for head size 512, but no code, issue or pull request that stores the Gemma~4 global KV cache once.

\end{document}